# 3rd Place Solution for Google Universal Image Embedding

Nobuaki Aoki
Kawasaki Heavy Industries, Ltd.

Yasumasa Namba
Ferix Inc.

kodoclanna@gmail.com

namba.nagoya@gmail.com

## Abstract

*This paper presents the 3rd place solution to the Google Universal Image Embedding Competition on Kaggle.* [1] *We use ViT-H/14 from OpenCLIP for the backbone of ArcFace, and trained in 2 stage. 1st stage is done with freezed backbone, and 2nd stage is whole model training. We achieve 0.692 mean Precision @5 on private leaderboard. Code available at*
*https://github.com/YasumasaNamba/google-universal-image-embedding*

## 1.Introduction

Image representation is a essential building block of computer vision applications . Traditionally, research on image embedding learning has focused on domain-specific models. However, models that are applicable only to a specific domain need to be created each time, which incurs costs. Therefore, there is a need to develop a general-purpose embedding model that can be applied to all domains with high accuracy. In this paper, we will introduce our techniques used in the Google Universal Image Embedding held on Kaggle. We describe an accurate model, a good traning method, and a lightweight inference of the model. We propose a general-purpose embedding model that can be applied to all domains with good accuracy.

## 2.Our Methods

### 2.1.Backbone Network

We selected a CLIP ViT-H/14 model trained with the LAION-2B English subset of LAION-5B. [2] It was a very good single model performance. CLIP model is better than other CNN and vision transformer based model trained with ImageNet. Furthermore OpenCLIP is more powerful than the official openai/CLIP. In our research, we also tried CLIP ViT-G/14 model trained with the LAION-2B English subset of LAION-5B[3], CLIP ViT-L/14 model trained with the LAION-2B English subset of LAION-5B[4], Openai/clip-vit-large-patch14-336[5] and others, but ViT-H/14 model trained with the LAION-2B English subset of LAION-5B showed the best performance. So We selected it.

### 2.2.Proposed Head

This competition requires the embedding dimensionality should be no greater than 64. Only single dense layer after backbone did not perform very well for that condition. After some trial and error, we created a customized head as Figure 1. Our customized model is consists of 20 parallel dense and dropout layers, and added at the end. The dropout rate is different in each layer. The reason for this is that the model output is 1280 (20×64=1280) in tensorflow Vit-H. This is because it was thought that when evaluating using k-nearest-neighbors with a dense of 64, even if the sum is finally added to form dense64, the accuracy would be better than simply linear dense64. This model can improve estimation accuracy. And it showed really good results. We also tried a larger 2560 (64x40) and a smaller 640 (64x10)., but it didn't work.

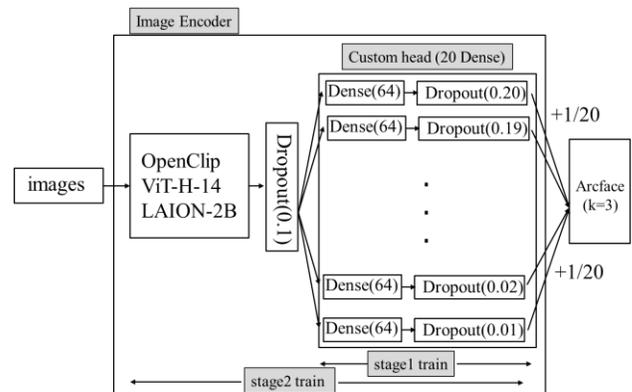

Figure 1: Proposed head.

### 2.3.Dataset and Augmentation

We use two public datasets prepared by @motono223, Google Landmark Recognition 2021[6] and Products-10K[7]. We tried to use other datasets, Alibaba goods dataset[8], The MET Dataset[9] and Large-scale

Fashion.(DeepFashion) Database[10], but they didn't work. Simple augmentation (fig. 2) is better in our experiment.

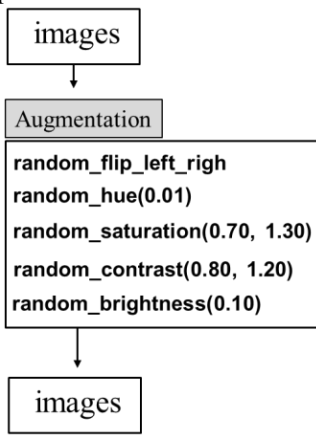

Figure 2: Simple augmentation.

### 2.4.Traning Strategy

Our method is a two-step training method, and our good scores are largely due to this learning method. We loaded the weights of the first stage and used them in the second stage.We used Arcface for learning. [11] First, freeze the backbone and train only the head at a high learning rate for 200 epochs(used step per epoch//10,So actually 20ep). Then stage2 trains for 10 epoch(used step per epoch//10,So actually 1ep) at a constant learning rate much lower than stage1.And, higher Arcface parameter m than stage1. By doing so, we could improve accuracy while suppressing overfitting. In our experiments, about 10 epochs gave the best scores without overfitting. I also tried 20 epochs and larger epochs, but the training loss decreased, but the Kaggle's Public Score decreased due to overfitting to the training data. It is presumed that this is because the types and number of test data are clearly larger than the traning data. For example the test data contains apparel and accessories,packaged,goods,landmarks,furniture&homedecor,storefronts,dishes,artwork,toys,memes,illustrations,cars. However, it is not enough to just increase the amount of data of that kind, but it is necessary to classify it with the same meaning as the test data, and it was quite difficult to find a good data set. Therefore, we performed training to suppress overfitting using only two datasets.

| | traning epoch | Arcface parameter | Training Lerning rate | Freeze | Environmental |
|---|---|---|---|---|---|
| First Stage | 200epoch (used step per epoch//10, So actually 20ep) | s=30 m=0.3 k=3 | Figure 2 | Backbone (OpenClipViT-H14) freeze | TPU |
| Second Stage | 10epoch (used step per epoch//10, So actually 1ep) | s=30 m=0.5 k=3 | Constant 1e-6 | No freeze | GPU |

Table 1: Training Parameter

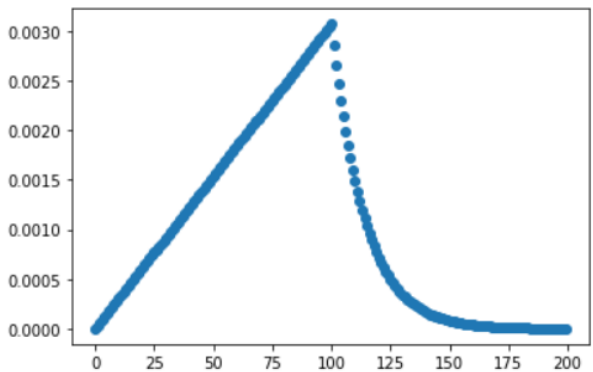

Figure 3: Stage1 Traning Lerning Rate

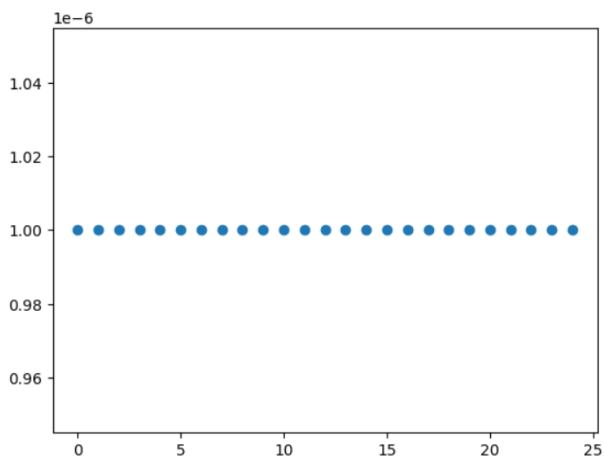

Figure 4: Stage2 Traning Lerning Rate

### 2.5.Evaluation method

It is difficult to create an evaluation metric when there are obviously few datasets that can be learned in the huge cases that have to be predicted. So we used Google Universal Image Embedding Competition on Kaggle's Public Leaderboard score. And if the score is good, we judge it as a success and work on improving the model.

### 2.6.Inferense with Pytorch

In this competition, submission with GPU or CPU is required, but when using tensorflow, there was a problem that a large model of Vit-H or higher consumes a lot of memory.Tensorflow can train quickly, but may use a lot of memory during inference. Therefore, weight reduction of the model can be achieved by extracting the weights once learned with tensorflow and porting them to pytorch.

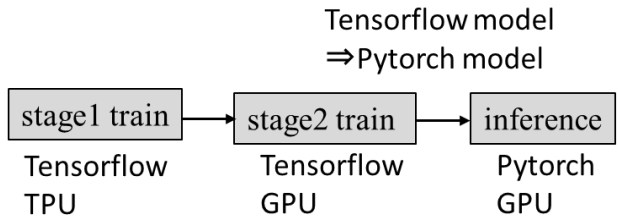

Figure 5: Environment.

## 3.Conclusion

We described a model with good accuracy, a good traning method, and a lightweight inference of the model. By using this method, good image representation becomes . We proposed a general-purpose embedding model that can be applied to all domains with good accuracy.